\documentclass[10pt,twocolumn,letterpaper]{article}

\usepackage[pagenumbers]{cvpr} 

\usepackage{graphicx}
\usepackage{amsmath}
\usepackage{amssymb}
\usepackage{booktabs}
\usepackage{threeparttable}
\usepackage{float}
\usepackage{verbatim}
\usepackage[dvipsnames]{xcolor}
\usepackage{comment}
\usepackage{amsmath}  
\usepackage{algorithm}
\usepackage{algorithmic}
\usepackage{colortbl}
\usepackage[colorlinks,linkcolor=blue]{hyperref}
\usepackage{diagbox}
\usepackage[accsupp]{axessibility}  
\definecolor{darkseagreen}{rgb}{0.56, 0.74, 0.56}
\definecolor{lightpink}{rgb}{1.0, 0.71, 0.76}

\usepackage[capitalize]{cleveref}
\crefname{section}{Sec.}{Secs.}
\Crefname{section}{Section}{Sections}
\Crefname{table}{Table}{Tables}
\crefname{table}{Tab.}{Tabs.}

\def\cvprPaperID{3422} 
\def\confName{CVPR}
\def\confYear{2022}

\begin{document}

\title{FocalClick: Towards Practical Interactive Image Segmentation }
\author{Xi Chen$^{1}$\quad Zhiyan Zhao$^{1}$\quad Yilei Zhang$^{1}$\quad Manni Duan$^{1}$\quad Donglian Qi$^{2}$\quad Hengshuang Zhao$^{3,4}$\\
$^{1}$Alibaba Group\quad $^{2}$Zhejiang University\quad $^{3}$The University of Hong Kong\\
$^{4}$Massachusetts Institute of Technology\\
}
\maketitle

\begin{abstract}
Interactive segmentation allows users to extract target masks by making positive/negative clicks. Although explored by many previous works, there is still a gap between academic approaches and industrial needs: first, existing models are not efficient enough to work on low-power devices;  second, they perform poorly when used to refine preexisting masks as they could not avoid destroying the correct part.
\textbf{FocalClick} solves both issues at once by predicting and updating the mask in localized areas. 
For higher efficiency, we decompose the slow prediction on the entire image into two fast inferences on small crops: a coarse segmentation on the Target Crop, and a local refinement on the Focus Crop.  To make the model work with preexisting masks, we formulate a sub-task termed Interactive Mask Correction, and propose Progressive Merge as the solution. Progressive Merge exploits morphological information to decide where to preserve and where to update, enabling users to refine any preexisting mask effectively.
FocalClick achieves competitive results against SOTA methods with significantly smaller FLOPs. It also shows significant superiority when making corrections on preexisting masks.
Code and data will be released at \href{https://github.com/XavierCHEN34/ClickSEG/}{github.com/XavierCHEN34/ClickSEG} 
\end{abstract}

\section{Introduction}
\begin{figure}[t]
\newcommand{\image}{\includegraphics[width=1\columnwidth]}
\centering 
\image{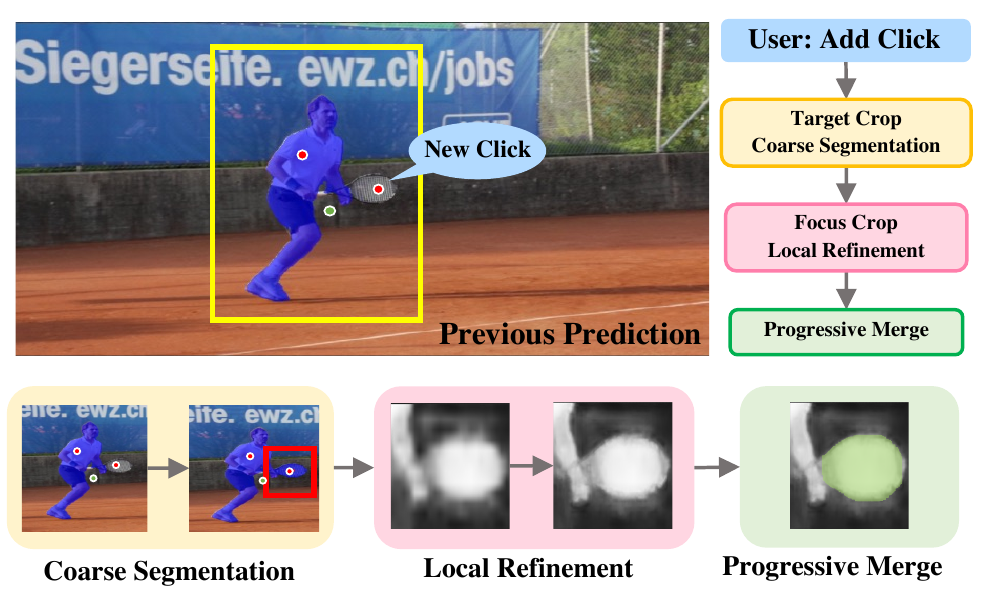} 
\vspace{-5mm}
\caption{
 After receiving a new click, FocalClick first selects a Target Crop~(yellow box) based on the previous mask to execute coarse segmentation. Then, we pick a Focus Crop~(red box) to make refinement according to the maximum different region between the coarse prediction and the previous mask. At last, Progressive Merge selects part of the new prediction to update.
}
\label{fig:1}
\vspace{-5mm}
\end{figure}

In recent years, interactive segmentation has aroused interest from both academia and industry. It enables users to annotate masks conveniently using simple interactions like scribbles~\cite{li2004lazy,grady2006random,bai2014error}, boxes~\cite{rother2004grabcut,wu2014milcut,lempitsky2009image}, or clicks~\cite{fbrs,firstclick,xu2016deep,jang2019brs,chen2021cdnet,sofiiuk2021ritm}. In this work, we focus on the click-based method. Under this setting, the users successively place positive/negative clicks~(like the red/green points in Fig.~\ref{fig:1}) to define the foreground and background, and the model returns new predictions after each user click.

The basic paradigm~\cite{xu2016deep} for click-based interactive segmentation is using Gaussian maps or disks to represent the clicks, then concatenating the click maps with the image as input, and sending it to a segmentation model to predict the mask. Based on this paradigm, previous works make improvements from different angles and keep improving SOTA. However, when applying them to practical scenarios, we find them not satisfactory in the following aspects.

\textbf{Efficiency on low-power devices.~}A good annotation tool is expected to produce fine masks with quick responses. Most previous works only focus on accuracy and take advantage of big models and high-resolution inputs.  However, they struggle when deployed on personal laptops, edge devices, or web service applications with high volume requests. When we attempt to reduce the input size for faster turnover, their accuracy drops significantly. 
       
\begin{figure}[t]
\newcommand{\image}{\includegraphics[width=0.48\columnwidth]}
\centering 
\tabcolsep=0.05cm
\renewcommand{\arraystretch}{0.06}
\begin{tabular}{cc}
\vspace{1mm}
\image{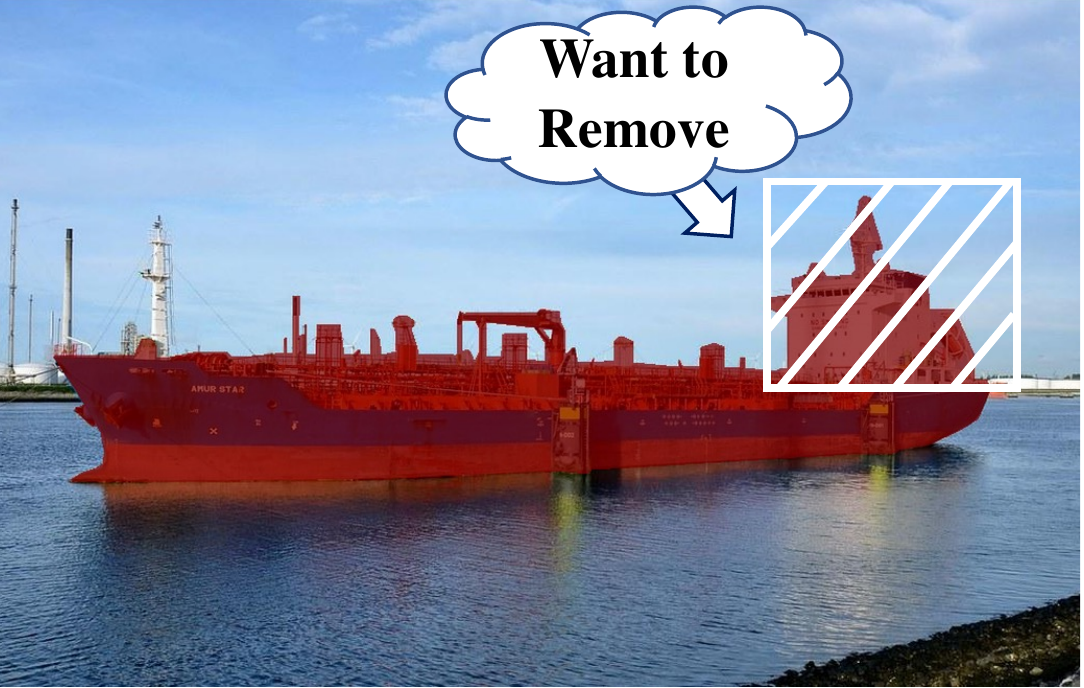} &
\image{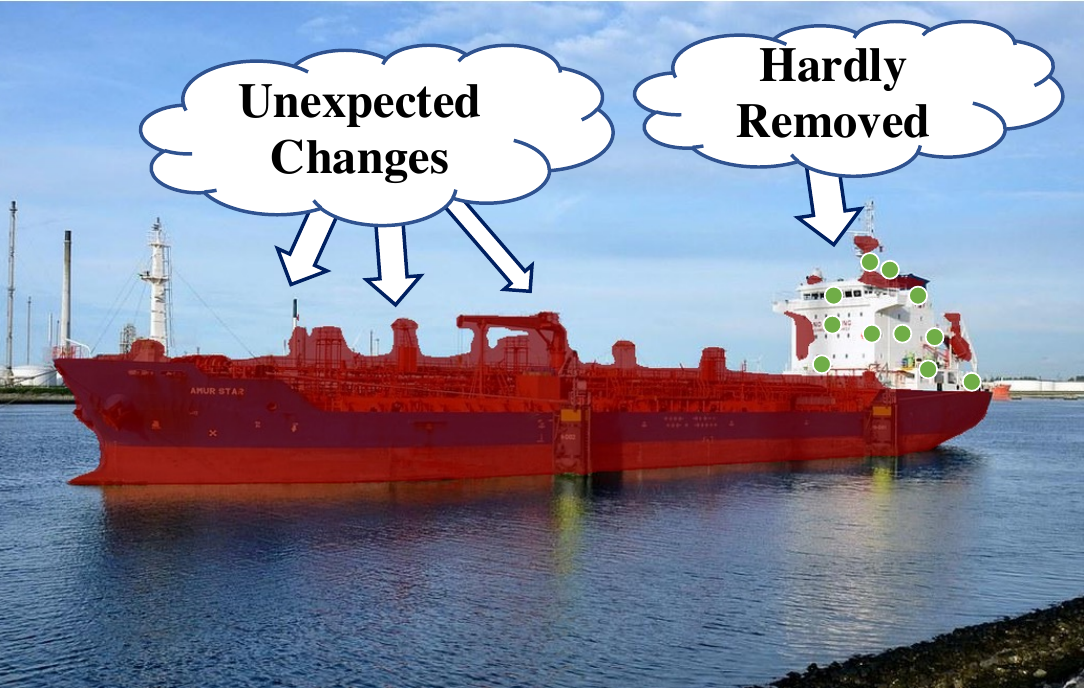} \\
\vspace{3mm}
{\footnotesize (a)~ Initial Mask } & {\footnotesize (b)~After Editing} \\
\end{tabular}
\vspace{-5mm}
\caption{Difficulties of using RITM\cite{sofiiuk2021ritm} to modify an initial mask. The negative clicks are marked in green in (b).}
\label{fig:failure}
\end{figure}

\textbf{Working with preexisting mask.~}In practical applications, there are not many mask annotation tasks that need to start from scratch. Preexisting masks are often provided by offline models or other forms of pre-processing. Making modifications on them could facilitate the annotation.
Nevertheless, previous methods are not compatible with externally given masks. 
Most works~\cite{fbrs,jang2019brs,firstclick,xu2016deep,chen2021cdnet,li2018latentdiversity} do not consider the previous mask as input.
Although \cite{sofiiuk2021ritm,forte2020getting99} concatenate the previous mask with the input image and the click maps, they do not perform well. As shown in Fig.~\ref{fig:failure}, RITM~\cite{sofiiuk2021ritm} hardly removes the mask of the cabin with several negative clicks. Moreover, it causes unexpected changes that are far away from the user clicks.

Facing those issues, we give an in-depth analysis of the existing pipeline. We notice that, during annotation, as the clicks are added successively by observing the current mask, each click has a specific target. For example, the target of the new click in Fig.~\ref{fig:1} is adding the racquet into the foreground. In this case, the racquet region requires more attention than the player and the tennis court.
However, previous works ignore this intention and make new predictions equally for all pixels after each click, which causes the two aforementioned issues: first, the well-predicted region would be computed repeatedly, causing computation redundancy; second, all pixels of the previous mask would be updated, making the model fail to preserve the details given in the initial masks.

With the above analysis in mind, we propose FocalClick to make the model focus on noteworthy patches. To get higher efficiency, FocalClick only makes predictions on the patch that really needs re-calculation. To make localized corrections on preexisting masks, FocalClick only updates the mask in the region that the user intends to modify and retains predictions in other regions.   

\begin{itemize}
    \item \textbf{ Efficient pipeline.~ }As in Fig.~\ref{fig:1},  given a new click, we first select a Target Crop according to the existing mask. This Target Crop would be resized into low resolution and execute coarse segmentation. Afterward, we localize a small region that the user intends to modify and send it into Refiner to extract the details. Thus, we decompose the time-consuming full image inference into two fast local predictions.

    \item \textbf{ Interactive mask correction.~}We formulate a new task termed Interactive Mask Correction and construct a benchmark to evaluate the ability to modify preexisting masks. We also propose Progressive Merge as the solution. Progressive Merge carries out morphological analysis of the previous masks and the current predictions to decide where to update and preserve. Thus, the correct part would not be destroyed.  
\end{itemize}

In general, our contributions could be summarized as follows: 1)~FocalClick is the first pipeline that deals with interactive segmentation from a completely local view, which reaches SOTA with significantly smaller FLOPs.  2)~We introduce the first benchmark and sub-task to evaluate the ability to correct preexisting masks, and propose corresponding solutions.  3)~This work makes interactive segmentation better meet the practical needs, which would be beneficial for both the industrial and academic world.

\section{Related Work}

\noindent \textbf{Interactive segmentation methods.~} Before the era of deep learning, researchers~\cite{rother2004grabcut,gulshan2010geodesic,grady2006random,kim2010nonparametric} take interactive segmentation as an optimization procedure.
DIOS~\cite{xu2016deep} first introduces deep learning into interactive segmentation by embedding positive and negative clicks into distance maps, and concatenating them with the original image as input. It formulates the primary pipeline and train/val protocol for click-based interactive segmentation. After this, \cite{li2018latentdiversity, liew2019multiseg} focus on the issue of ambiguity and predict multiple potential results and let a selection network or the user choose from them. FCANet~\cite{firstclick} emphasizes the particularity of the first click and uses it to construct visual attention. BRS~\cite{jang2019brs} first introduces online optimization, which enables the model to update during annotation. f-BRS~\cite{fbrs} speeds up the BRS~\cite{jang2019brs} by executing online optimization in specific layers.
CDNet~\cite{chen2021cdnet} introduces self-attention into interactive segmentation to predict more consistent results. RITM~\cite{sofiiuk2021ritm} and 99\%AccuracyNet~\cite{forte2020getting99} add the previous mask as network input to make the prediction more robust and accurate. 
These methods achieve excellent performances, but they suffer from slow inference speed, and they could not deal with preexisting masks.\\

\noindent \textbf{ Local inference for interactive segmentation.~}Some previous methods~\cite{fbrs,chen2021cdnet,sofiiuk2021ritm} also crop the region around the last predicted target for the following step inference, which is similar to our Target Crop. However, as they predict the final masks on this crop, they require to keep the resolution. Differently, FocalClick leverages the Target Crop to locate the Focus Crop area, and does not rely on the Segmentor to produce fine details, enabling us to resize the Target Crop to small scales for higher speed. 

\cite{liew2017regional,hao2021edgeflow,forte2020getting99} follow the similar coarse to fine schema and add additional modules to refine the primitive predictions. Nevertheless, RIS-Net~\cite{liew2017regional} crop multiple rois only based on click positions to make the refinement. EdgeFlow~\cite{hao2021edgeflow} and 99\%AccuracyNet \cite{forte2020getting99} make refinement for the boundary of the primitive predictions. They all get finer results at the cost of larger computations. In contrast, FocalClick picks the local patches with more focusing strategies. It significantly reduces the FLOPs by decomposing the pipeline into coarse segmentation and refinement, thus achieving our goal of making interactive segmentation more practical.

\begin{figure*}[t]
\newcommand{\image}{\includegraphics[width=2\columnwidth]}
\centering 
\image{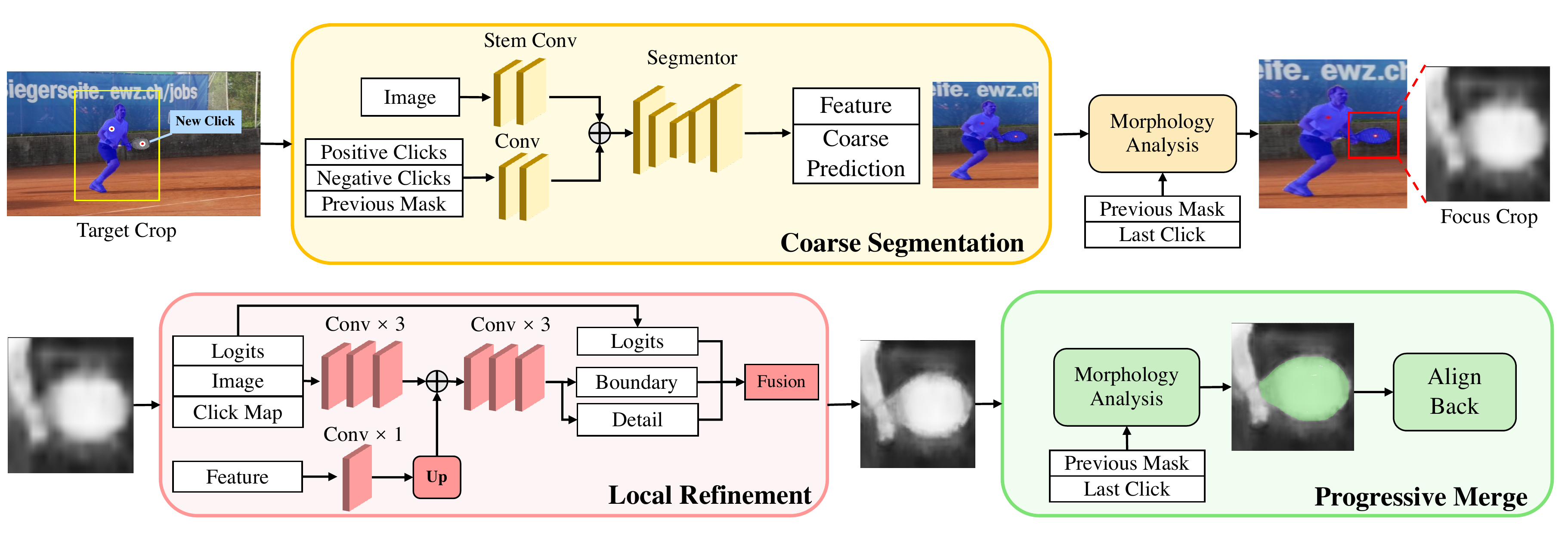} 
\vspace{-3mm}
\caption{ We take the image, two click maps, and the previous mask as input. We use binary disks with radius 2 to represent the click. First, we select the \textbf{Target Crop} around the target object and resize it to a small size. It is then sent into \textbf{Segmentor} to predict a coarse mask. Next, we chose a \textbf{Focus Crop} by calculating the different regions between the previous masks and the coarse prediction to refine the details. At last,  \textbf{Progressive Merge} updates the local part that the user intends to modify and preserves the details in other regions.
}
\label{fig:pipeline}
\vspace{-3mm}
\end{figure*}

\section{Method}

We propose FocalClick to bridge the gap between existing methods and industrial needs. We first introduce our efficient pipeline, then elaborate on a new task termed Interactive Mask Correction and our benchmark.

\subsection{Efficient Pipeline}
The key of our pipeline is to decompose one heavy inference on the whole image into two light predictions on small patches.
The pipeline of FocalClick is demonstrated in Fig.~\ref{fig:pipeline}.  First, Target Crop chooses the patch around the target object, resizes it to a small scale, and sends it to Segmentor to predict a coarse mask. Afterward, Focus Crop picks a local region that needs refinement and feeds the zoom-in local patch into Refiner. At last, Progressive Merge aligns the local predictions back to the full-scale masks. Thus, we only refine a small local region after each click, but all pixels of the final prediction have been refined with the calculation allocated to different rounds.\\

\noindent \textbf{Target crop.~}The objective is to filter out background information that is not related to the target object. We first calculate the minimum external box of the previous mask and the newly added click, and expand it with a ratio $r_{TC}=1.4$ following \cite{fbrs,sofiiuk2021ritm}. Then, we crop the input tensors~(image, previous mask, click maps) and resize them into small scales. \\ 

\noindent \textbf{Coarse segmentation.~}We expect to get a rough mask for the target, which could assist us in locating the Focus Crop and make a further refinement. The Segmentor could be any segmentation network~\cite{pspnet,chen2017deeplab,long2015fcn,xiao2018upernet,peng2017largekernel} for customized scenarios. We select the current SOTA methods HRNet+OCR~\cite{wang2020hrnet,yuan2020ocr} and SegFormer~\cite{xie2021segformer} as representatives. As demonstrated in Fig.~\ref{fig:pipeline}, we follow RITM~\cite{sofiiuk2021ritm} to use two conv layers to adjust the channel and scale of the click maps and make feature fusion after the stem layers. \\

\noindent \textbf{Focus crop.~}It aims to locate the area that the user intends to modify. We first compare the differences between the primitive segmentation results and the previous mask to get a Difference Mask $ {M_{xor}}$. We then calculate the max connected region of ${M_{xor}}$ that contains the new click, and we generate the external box for this max connected region.
Similar to the Target Crop, we make expansion with ratio $r_{FC}=1.4$. We note this region Focus Crop. Accordingly, we crop local patches on the input image and click maps. Besides, we use RoiAlign~\cite{he2017mask} to crop the feature and the output logits predicted by the Segmentor. \\

\noindent \textbf{Local refinement.~}It recovers the details of the coarse prediction in Focus Crop. 
We first extract the low-level feature from the cropped tensor using Xception convs~\cite{chollet2017xception}. At the same time, we adjust the channel number of the RoiAligned feature and fuse it with the extracted low-level feature. To get refined predictions, we utilize two heads to predict a Detail Map $M_d$ and a Boundary Map $M_b$, and calculate the refined prediction $M_r$ by updating the boundary region of the coarse predicted logits $M_l$, as in Eq.~\ref{eq:refine}.
\begin{equation}
\small
\mathit{M_r  =  Sigmoid(M_b) * M_d + (1 - Sigmoid(M_b)) * M_l}
\label{eq:refine}
\end{equation}

\noindent \textbf{Progressive merge.~}When annotating or editing masks, we do not expect the model to update the mask for all pixels after each click. Otherwise, the well-annotated details would be completely over-written. Instead, we only want to update in limited areas that we intend to modify.  
Similar to the method of calculating Focus Crop, Progressive Merge distinguishes the user intention using morphology analysis. After adding a user click, we simply binarize the newly predicted mask with a threshold of 0.5 and calculate a different region between the new prediction and the preexisting mask, then pick the max connected region that contains the new click as the update region~(the part the green in Fig.~\ref{fig:pipeline}).  In this region, we update the newly predicted mask onto the previous mask, and we keep the previous mask untouched in other regions.  

When starting with a preexisting mask or switching back from other segmentation tools, we apply Progressive Merge to preserve the correct details. While annotating from scratch, we active the progressive mode after 10 clicks.\\

\noindent \textbf{Training supervision.~}The supervision of the boundary map $M_b$ is computed by down-sampling the segmentation ground truth 8 times and resizing it back. The changed pixels could represent the region that needs more details. We supervise the boundary head with Binary Cross Entropy Loss~${L_{bce}}$. The coarse segmentation is supervised by Normalized Focal Loss~$L_{nfl}$ proposed in RITM~\cite{sofiiuk2021ritm}. For the refined prediction, we add boundary weight~(1.5) on the NFL loss, and we note it as $L_{bnfl}$, the total loss could be calculated as Eq.~\ref{eq:loss}.
\begin{equation}
\mathit{L = L_{bce} + L_{nfl} + L_{bnfl} }
\label{eq:loss}
\end{equation}

\subsection{ Interactive Mask Correction }
\label{sec:mask}
In practical applications, large portions of annotation tasks provide pre-inferred masks. In this case, annotators only need to make corrections on them instead of starting from zero. Besides, during annotation, when annotators switch to matting, lasso, or polygon tools and switchback, we also expect to preserve the pixels annotated by other tools. However, existing methods predict new values for all pixels after each click. Thus, they are not compatible with modifying preexisting masks or incorporating other tools.

To solve this problem, we make the following attempts:
1)~We construct a new benchmark, DAVIS-585, which provides initial masks to measure the ability of mask correction.
2)~We prove that our FocalClick shows significant superiority over other works in this new task.\\

\noindent \textbf{New benchmark:~DAVIS-585. }Existing works uses GrabCut~\cite{rother2004grabcut}, Berkeley~\cite{berkeley}, DAVIS~\cite{davis}, SBD~\cite{SBD} to evaluate the performance of click-based interactive segmentation. However, none of them provides initial masks to measure the ability of Interactive Mask Correction. 
Besides, GrabCut and Berkeley only contain 50 and 100 easy examples, making the results not convincing. SBD provides 2802 test images, but they are annotated by polygons with low quality. DAVIS is firstly introduced into interactive segmentation in \cite{li2018latentdiversity}. It contains 345 high-quality masks for diverse scenarios. However, as \cite{li2018latentdiversity} follows the setting of DAVIS2016, it merges all objects into one mask. Hence, it does not contain small objects, occluded objects, and no-salient objects. In this paper, we chose to build a new test set based on DAVIS~\cite{davis} for its high annotation quality and diversity, and we made two modifications:

First, we follow DAVIS2017 which annotates each object or accessory separately, making this dataset more challenging.  We uniformly sample 10 images per video for 30 validation videos, and we take different object annotations as independent samples. Then we filter out the masks under 300 pixels and finally get 585 test samples, so we call our new benchmark \textbf{DAVIS-585}.

Second, to generate the flawed initial masks, we compare two strategies:~1)~Simulating the defects on ground truth masks using super-pixels. 2)~Generating defective masks using offline models. We find the first strategy has two advantages: 1)~It could control the distribution of error type and the initial IOUs. 2)~The simulated masks could be used to measure the ability to preserve the correct part of preexisting masks. 
Therefore, we use super-pixels algorithm\footnote{https://github.com/Algy/fast-slic} to simulate the defect.  We first use mask erosion and dilation to extract the boundary region of the ground truth mask. We then define three types of defects: boundary error, external FP~(False positive), and internal TN~(True Negative).
After observing the error distribution in real tasks,  we set the probability of these three error types to be [0.65, 0.25, 0.1] and follow Alg.~\ref{alg1} to control the quality of each defective mask. To decide the quality range, we carry out a user study and find that users intend to discard the given masks when they have IOU lower than 75\%. Considering that current benchmarks use NoC85~(Number of Clicks required on average to reach IOU 85\%) as the metric, we control our simulated masks to have IOUs between 75\% and 85\%. \\

\begin{algorithm}[t]
    \small
	\caption{Simulate Defective Mask using Super-Pixels} 
	\label{alg1} 
	\small
	\begin{algorithmic}
		\REQUIRE Image, GTMask, maxIOU=0.85, minIOU=0.75
		\STATE SimMask $\gets$ GT
		\WHILE{ $\mathrm{True} $ } 
		\STATE ErrorType $\gets$ Rand([Boundary, External, Internal])
		\STATE PixelNumber $\gets$ Rand([50, 100, 200, 300, 500, 700])
		\STATE SuperPixels $\gets$ Slic(Image, PixelNumber)
		\STATE SimMask $\gets$ Merge(~ErrorType, SuperPixels, SimMask)
		\STATE MaskIOU $\gets$ IOU(~SimMask, GTMask)
		
		\IF{MaskIOU $<$ minIOU} 
		\STATE Break
		\ELSIF{MaskIOU $>$ maxIOU}
		\STATE Continue 
		\ELSE
		\STATE Return SimMask
		\ENDIF 
		\ENDWHILE 
	\end{algorithmic}
\end{algorithm}
\vspace{-3mm}

\section{Experiment}

We first introduce the basic settings for our models and the train/val protocols for click-based interactive segmentation. Then, we compare FocalClick with current SOTA methods on existing benchmarks for both accuracy and efficiency.  Next, we report the performance on our new benchmark DAVIS-585. Afterward, we conduct ablation studies to verify the effectiveness of our method for both interactive segmentation and mask correction.

\subsection{Experimental Configuration}
\noindent \textbf{Model series.~}To satisfy the requirement for different scenarios, we design two versions of models as demonstrated in Table~\ref{tab:series}. The S1 version is adapted to edge devices and the plugins for web browsers.  The S2 version would be suitable for CPU laptops. In this paper, we conduct experiments for both SegFormer~\cite{xie2021segformer} and HRNet~\cite{wang2020hrnet} as our Segmentor to show the universality of our pipeline. In the rest of the paper, we use S1, S2 to denote different versions of our model. 
\begin{table}[h]
\small
\begin{center}
\begin{threeparttable}
\begin{tabular}{l|c|c  }
\toprule[1pt]
Model Series       & Segmentor Input &  Refiner Input \\
\hline
FocalClick-S1  & $128\times128$ & $256\times256$ \\
FocalClick-S2  & $256\times256$ & $256\times256$ \\
\bottomrule[1pt]
\end{tabular}
\end{threeparttable}
\end{center}
\vspace{-5mm}
\caption{ Configurations of FocalClick series.  
}
\label{tab:series}
\end{table}

\noindent \textbf{Training protocol.~}We simulate the Target Crop by randomly cropping a region with a size of $256\times256$ on the original images. Then, we simulate the Focus Crop by calculating the external box for the ground truth mask, or making random local crops centered on boundaries with the length of 0.2 to 0.5 of the object length. 
Then, we add augmentations of random expansion from 1.1 to 2.0 on the simulated Focus Crop. Thus, the whole pipeline of Segmentor and Refiner is trained together in an end-to-end manner.

For the strategy of click simulation, we exploit iterative training~\cite{mahadevan2018iteratively} following RITM~\cite{sofiiuk2021ritm}. Besides the iteratively added clicks, the initial clicks are samples inside/outside the ground truth masks randomly following \cite{xu2016deep}. The maximum number of positive~/negative clicks is set as 24 with a probability decay of 0.8. 

For the hyper-parameters, following RITM~\cite{sofiiuk2021ritm}, we train our models on a combination dataset of COCO~\cite{lin2014coco} and LVIS~\cite{gupta2019lvis}. We also report the performance of our model trained on SBD~\cite{SBD}, and a large combined dataset~\cite{lin2014coco,gupta2019lvis,borji2015msra10k,ade20k,wang2017learningduts,xu2018youtubevos,thinobject,cong2020dovenet}.
During training, we only use flip and random resize with the scale from 0.75 to 1.4 as data augmentation.  We apply Adam optimizer of $\beta_1 = 0.9, \beta_1 = 0.999$. We denote 30000 images as an epoch and train our models with 230 epochs. We use the initial learning rate as $5\times 10^{-4}$ and decrease it 10 times at the epoch of 190 and 220. We train each of our models on two V100 GPUs with batch size 32. The training takes around 24 hours. \\

\noindent \textbf{Evaluation protocol.~}We follow previous works~\cite{xu2016deep,chen2021cdnet,sofiiuk2021ritm,jang2019brs,fbrs,firstclick} to make fair comparisons. During evaluation, the clicks are automatically simulated with a fixed strategy:  Each click would be placed at the center of the largest error region between the previous prediction and the ground truth mask. For example, when starting from scratch, the first click would be placed at the center of the ground truth mask. Additional clicks would be added iteratively until the prediction reaches the target IOU~(Intersection over Union) or the click number reaches the upper bound.    

For the metrics, we report NoC~IOU~(Number of Clicks), which means the average click number required to reach the target IOU. Following previous works, the default upper bound for click number is 20. We note the sample as a failure if the model fails to reach the target IOU within 20 clicks. Hence, we also report NoF~IOU~(Numbers of Failures) to measure the average number of failures.

\begin{table*}[t]
\begin{center}
\scalebox{0.8}{
\begin{tabular}{ll|c|c|c|c|c|c|c|c}
\toprule[1pt]
\multicolumn{3}{l|}{} & \multicolumn{2}{c|}{GrabCut~\cite{rother2004grabcut}} & Berkeley~\cite{berkeley} & \multicolumn{2}{c|}{SBD~\cite{SBD}} & \multicolumn{2}{c}{DAVIS~\cite{davis}} \\
\cline{4-10}
\multicolumn{2}{l|}{Method } & Train Data   & NoC~85 & NoC~90 & NoC~90 & NoC~85 & NoC~90 & NoC~85 & NoC~90 \\
\hline
\multicolumn{2}{l|}{Graph cut~\cite{boykov2001interactive}} & /  & 7.98 & 10.00 & 14.22 & 13.6 & 15.96 & 15.13 & 17.41 \\
\multicolumn{2}{l|}{Geodesic matting~\cite{gulshan2010geodesic}} & / & 13.32 & 14.57 & 15.96 & 15.36 & 17.60 & 18.59 & 19.50 \\
\multicolumn{2}{l|}{Random walker~\cite{grady2006random}}  & / & 11.36 & 13.77 & 14.02 & 12.22 & 15.04 & 16.71 & 18.31 \\
\multicolumn{2}{l|}{Euclidean star convexity~\cite{gulshan2010geodesic}}  & /  & 7.24 & 9.20 & 12.11 & 12.21 & 14.86 & 15.41 & 17.70 \\
\multicolumn{2}{l|}{Geodesic star convexity~\cite{gulshan2010geodesic}}  & / & 7.10 & 9.12 & 12.57 & 12.69 & 15.31 & 15.35 & 17.52 \\
\hline
\multicolumn{2}{l|}{DOS w/o GC~\cite{xu2016deep}}  & \scalebox{0.75}{ Augmented VOC\cite{everingham2010pascal,SBD} } & 8.02 & 12.59 & -- & 14.30 & 16.79 & 12.52 & 17.11 \\
\multicolumn{2}{l|}{DOS with GC~\cite{xu2016deep}}  & \scalebox{0.75}{ Augmented VOC\cite{everingham2010pascal,SBD} } & 5.08 & 6.08 & -- & 9.22 & 12.80 & 9.03 & 12.58 \\
\multicolumn{2}{l|}{RIS-Net~\cite{liew2017regional}}  & \scalebox{0.75}{ Augmented VOC\cite{everingham2010pascal,SBD} }& -- & 5.00 & -- & 6.03 & -- & -- & -- \\
\multicolumn{2}{l|}{CM guidance~\cite{majumder2019content}}  & \scalebox{0.75}{ Augmented VOC\cite{everingham2010pascal,SBD} }& -- & 3.58 & 5.60 & -- & -- & -- & --\\
\multicolumn{2}{l|}{FCANet~(SIS)~\cite{firstclick}}  & \scalebox{0.75}{ Augmented VOC\cite{everingham2010pascal,SBD} } & -    & 2.14 & 4.19 &-     & -    & -    & 7.90 \\
\multicolumn{2}{l|}{Latent diversity~\cite{li2018latentdiversity}}  & \scalebox{0.75}{SBD\cite{SBD}} & 3.20 & 4.79 & -- & 7.41 & 10.78 & 5.05  & 9.57 \\
\multicolumn{2}{l|}{BRS~\cite{jang2019brs}}  & \scalebox{0.75}{SBD\cite{SBD}} & 2.60 & 3.60 & 5.08 & 6.59 & 9.78 & 5.58 & 8.24 \\
\multicolumn{2}{l|}{f-BRS-B-resnet50~\cite{fbrs}}  & \scalebox{0.75}{SBD\cite{SBD}} & 2.50 & 2.98 & {4.34} & 5.06 & 8.08 & 5.39 & 7.81 \\
\multicolumn{2}{l|}{ CDNet-resnet50~\cite{chen2021cdnet}}  & \scalebox{0.75}{SBD\cite{SBD}} &  2.22 & 2.64 & 3.69 & 4.37 &  7.87& {5.17} & 6.66 \\
\multicolumn{2}{l|}{ RITM-hrnet18~\cite{sofiiuk2021ritm}}  & \scalebox{0.75}{SBD\cite{SBD}} &  1.76 & 2.04 & 3.22 & \textbf{3.39} &  \textbf{5.43} & {4.94} & 6.71 \\
\rowcolor{gray!20} 
\multicolumn{2}{l|}{ Ours-hrnet18s-S2 } &  \scalebox{0.75}{SBD\cite{SBD}} & 1.86  & 2.06 & \textbf{3.14}  & 4.30 &  6.52 & \textbf{4.92} & \textbf{6.48}  \\
\rowcolor{gray!20} 
\multicolumn{2}{l|}{ Ours-segformerB0-S2}  & \scalebox{0.75}{SBD\cite{SBD}} & \textbf{1.66}   &  \textbf{1.90} & {3.14}  & 4.34  & 6.51 & 5.02  & 7.06 \\
\hline
\multicolumn{2}{l|}{ 99\%AccuracyNet~\cite{forte2020getting99}}  & \scalebox{0.75}{Synthetic\cite{SBD,lin2014coco,xu2017dim,dai2014synthesizability}}   &  - & 1.80 & 3.04 & 3.90 &  - & - & - \\
\multicolumn{2}{l|}{f-BRS-B-hrnet32~\cite{fbrs}}  & \scalebox{0.75}{ COCO\cite{lin2014coco}+LVIS\cite{gupta2019lvis}}  & 1.54 & 1.69 & 2.44 & 4.37 & 7.26 & 5.17 & 6.50 \\
\multicolumn{2}{l|}{ RITM-hrnet18s~\cite{sofiiuk2021ritm}}  & \scalebox{0.75}{ COCO\cite{lin2014coco}+LVIS\cite{gupta2019lvis}} &  1.54 & 1.68 & 2.60 & 4.04 &  6.48 & 4.70 & 5.98 \\
\multicolumn{2}{l|}{ RITM-hrnet32~\cite{sofiiuk2021ritm}}  & \scalebox{0.75}{ COCO\cite{lin2014coco}+LVIS\cite{gupta2019lvis}} &  1.46 & 1.56 & 2.10 & 3.59 & 5.71 & 4.11 & 5.34 \\
\multicolumn{2}{l|}{ EdgeFlow-hrnet18~\cite{hao2021edgeflow}}  & \scalebox{0.75}{ COCO\cite{lin2014coco}+LVIS\cite{gupta2019lvis}} &  1.60 & 1.72 & 2.40 & - &  - & 4.54 & 5.77 \\
\rowcolor{gray!20} 
\multicolumn{2}{l|}{ Ours-hrnet18s-S1 } &  \scalebox{0.75}{ COCO\cite{lin2014coco}+LVIS\cite{gupta2019lvis}}  &  1.64 & 1.82 & 2.89 & 4.74 & 7.29 & 4.77 & 6.56 \\
\rowcolor{gray!20} 
\multicolumn{2}{l|}{ Ours-hrnet18s-S2 } &  \scalebox{0.75}{ COCO\cite{lin2014coco}+LVIS\cite{gupta2019lvis}}   &  1.48 & 1.62 & 2.66 & 4.43 &  6.79 & 3.90 & 5.25 \\
\rowcolor{gray!20} 
\multicolumn{2}{l|}{ Ours-hrnet32-S2 } &  \scalebox{0.75}{ COCO\cite{lin2014coco}+LVIS\cite{gupta2019lvis}}   & 1.64  & 1.80  & 2.36 & 4.24 &  6.51 & 4.01 & 5.39 \\
\rowcolor{gray!20} 
\multicolumn{2}{l|}{ Ours-segformerB0-S1}  & \scalebox{0.75}{ COCO\cite{lin2014coco}+LVIS\cite{gupta2019lvis}} & 1.60  & 1.86  & 3.29  & 4.98 & 7.60  & 5.13  & 7.42 \\
\rowcolor{gray!20} 
\multicolumn{2}{l|}{ Ours-segformerB0-S2}  & \scalebox{0.75}{ COCO\cite{lin2014coco}+LVIS\cite{gupta2019lvis}}  &  \textbf{1.40} & 1.66 & 2.27 & 4.56 & 6.86 & 4.04 & 5.49 \\
\rowcolor{gray!20} 
\multicolumn{2}{l|}{ Ours-segformerB3-S2}  & \scalebox{0.75}{ COCO\cite{lin2014coco}+LVIS\cite{gupta2019lvis}}  & 1.44  &  \textbf{1.50} & \textbf{1.92}  & \textbf{3.53} & \textbf{5.59} & \textbf{3.61} & \textbf{4.90} \\
\hline
\rowcolor{gray!20} 
\multicolumn{2}{l|}{ Ours-hrnet32-S2 } &  \scalebox{0.75}{Large Dataset}   & 1.30  & 1.34 & 1.85 & 4.35 &  6.61 & 3.19 & 4.81 \\
\rowcolor{gray!20} 
\multicolumn{2}{l|}{ Ours-segformerB3-S2}  & \scalebox{0.75}{Large Dataset} &  1.22 & 1.26 & 1.48 & 3.70 &  5.84 & 2.92 & 4.52 \\

\bottomrule[1pt]
\end{tabular}
}
\end{center}
\caption{Evaluation results on GrabCut, Berkeley, SBD and DAVIS datasets. 
 `NoC~85/90' denotes the average Number of Clicks required the get IoU of 85/90\%. `Synthetic' data uses the datasets of \cite{lin2014coco,xu2017dim,dai2014synthesizability}.  `Large Dataset' denotes a combined dataset.~\cite{lin2014coco,gupta2019lvis,borji2015msra10k,ade20k,wang2017learningduts,xu2018youtubevos,thinobject,cong2020dovenet} }
\label{tab:evaluation sota}
\vspace{-2mm}
\end{table*}

\subsection{ Comparison with State-of-the-Art}

\noindent \textbf{Computation analysis.~}The objective of FocalClick is to propose a practical method for mask annotation; efficiency is a significant factor. In Table~\ref{tab:computation}, we make a detailed analysis and comparison for the number of parameters, FLOPs, and the inference speed on CPUs. We summarize the methods of the predecessors into five prototypes according to their backbone and input size.

\begin{table}[h]
\small
\begin{center}
\scalebox{0.8}{
\begin{tabular}{ll|c|c|c|c|c|c}
\toprule[1pt]
\multicolumn{2}{l|}{} & \multicolumn{2}{c|}{Params/MB} & \multicolumn{2}{c|}{FLOPs/G} & \multicolumn{2}{c}{Speed/ms}  \\
\cline{3-8}
\multicolumn{2}{l|}{Base Model } & Seg & Ref & Seg & Ref   & Seg & Ref \\
\hline
\multicolumn{2}{l|}{hrnet18s-400~\cite{sofiiuk2021ritm,hao2021edgeflow}} &  4.22 & 0 & 8.96 & 0 & 470 & 0  \\
\multicolumn{2}{l|}{hrnet18s-600~\cite{sofiiuk2021ritm,hao2021edgeflow}} &  4.22 & 0 & 20.17 & 0 & 1020 & 0  \\
\multicolumn{2}{l|}{hrnet32-400~\cite{sofiiuk2021ritm}} &  30.95 & 0 & 40.42 & 0 & 1387 & 0  \\
\multicolumn{2}{l|}{resnet50-400~\cite{chen2021cdnet,fbrs,forte2020getting99}} &  31.38 & 0 & 84.63 & 0 & 2359 & 0  \\
\multicolumn{2}{l|}{resnet101-512~\cite{firstclick}} &  50.37 & 0 & 216.55 & 0 & 6267 & 58 \\
\hline
\multicolumn{2}{l|}{Ours-B0-S1} &  3.72 & 0.016 & 0.43 & 0.17 & 41 & 59  \\
\multicolumn{2}{l|}{Ours-B0-S2} &  3.72 & 0.016 & 1.77 & 0.17 & 140 & 60  \\
\multicolumn{2}{l|}{Ours-B3-S2} &  45.6 & 0.025 & 12.72 & 0.20 & 634 & 72 \\
\multicolumn{2}{l|}{Ours-hrnet18s-S1} &  4.22 & 0.011 & 0.91 & 0.15 & 80 & 50  \\
\multicolumn{2}{l|}{Ours-hrnet18s-S2} &  4.22 & 0.011 & 3.66 & 0.16 & 213 & 51  \\
\multicolumn{2}{l|}{Ours-hrnet32-S2} & 30.95  & 0.025 & 16.92 & 0.20 & 650 & 51  \\
\bottomrule[1pt]
\end{tabular}
}
\end{center}
\vspace{-5mm}
\caption{Computation analysis for FocalClick Series and SOTA methods. `B0/3' is short for SegFormer-B0/3. `Seg' denotes Segmentor, `Ref' denotes Refiner. `400', `600', `512' denote the default input size required by different models. The speed is measured on a CPU laptop with 2.4 GHz, 4$\times$Intel Core i5.  }
\label{tab:computation}
\vspace{-5mm}
\end{table}

In Table~\ref{tab:computation}, most works use big models and 400 to 600 input sizes, which makes them hard to use on CPU devices. In contrast, FocalClick supports light models and small input sizes like 128 and 256. The FLOPs of our B0-S1 version is 15 times smaller than the lightest RITM~\cite{sofiiuk2021ritm}, 360 times smaller than  FCANet~\cite{firstclick}. Using the same Segmentor, our hrnet-18s version could reduce 2 to 8 times FLOPs compared with original RITM~\cite{sofiiuk2021ritm}. 

Besides, as FocalClick is a universal pipeline that could be adapted to various Segmentors, the FLOPs could be further reduced by using more light-weighted architectures like \cite{howard2017mobilenets,tan2019efficientnet}.\\

\noindent \textbf{Performance on existing benchmarks. }The comparison results on existing benchmarks are demonstrated in Table~\ref{tab:evaluation sota}.
We split previous methods into different blocks according to the training data. Some early works are trained on SBD~\cite{SBD} or Augmented VOC~\cite{SBD,everingham2010pascal} which are relatively small. Recently SOTA methods use COCO\cite{lin2014coco} and LVIS~\cite{gupta2019lvis} for training and getting better results. For a fair comparison, we report our performance under both training settings.
We note that, with significantly smaller FLOPs, different versions of FocalClick all demonstrate competitive or superior performance against previous SOTA methods.

As our goal is developing a practical method, at the bottom of Table~\ref{tab:evaluation sota}, to explore the upper-bound of FocalClick in the practical scenario, we also report results trained on a combined dataset~\cite{lin2014coco,gupta2019lvis,borji2015msra10k,ade20k,wang2017learningduts,xu2018youtubevos,thinobject,cong2020dovenet}. It shows that when equipped with large training data, FocalClick could outperform other methods with a large margin.

\begin{table*}[t]
\small
\begin{center}
\scalebox{0.85}{
\begin{tabular}{ll|c|c|c|c|c|c|c|c|c|c|c|c}
\toprule[1pt]
\multicolumn{2}{l|}{} & \multicolumn{6}{c|}{From Initial Mask} & \multicolumn{6}{c}{From Scratch}   \\
\cline{3-14}
\multicolumn{2}{l|}{Method } & NoC85 & NoC90 & NoC95 & NoF85 & NoF90 & NoF95 & NoC85 & NoC90 & NoC95 & NoF85 & NoF90 & NoF95 \\

\hline
\multicolumn{2}{l|}{RITM-hrnet18s~\cite{sofiiuk2021ritm}} &  3.71 & 5.96 & 11.83 & 49 & 80 & 235 &        5.34 & 7.57 & 12.94 & 52 & 91 & 257  \\
\multicolumn{2}{l|}{RITM-hrnet32~\cite{sofiiuk2021ritm}} &  3.68 & 5.57 & 11.35 & 46 & 75 & 214 &         4.74 & 6.74 & 12.09 & 45 & 80 & 230  \\
\hline
\multicolumn{2}{l|}{Ours-hrnet18s-S1} &  2.72 & 3.82 & 5.86 & 37 & 57 & 97 & 5.62 & 8.08 & 13.73 & 53 & 98 & 274 \\
\multicolumn{2}{l|}{Ours-hrnet18s-S2} & 2.48 & 3.34 & 5.18 & 31 & 43 & 79 & 4.93 & 6.87 & 11.97 & 49 & 77 & 239  \\
\multicolumn{2}{l|}{Ours-hrnet32-S2} & 2.32 & 3.09 & 4.94 & 28 & 41 & 74 & 4.77 & 6.84 & 11.90 & 48 & 76 & 241  \\
\multicolumn{2}{l|}{Ours-segformerB0-S1} &  2.63 & 3.69 & 6.08 & 38 & 54 & 104 &  6.21 & 9.06 & 14.81 & 64 & 127 & 315  \\
\multicolumn{2}{l|}{Ours-segformerB0-S2} &  2.20 & 3.08 & 4.82 & 27 & 39 & 68 & 4.99 & 7.13 & 12.65 & 50 & 86 & 260   \\
\multicolumn{2}{l|}{Ours-segformerB3-S2} &  2.00 & 2.76 & 4.30 & 22 & 35 & 53 & 4.06 & 5.89 & 11.12 & 43 & 74 & 218   \\
\bottomrule[1pt]
\end{tabular}
}
\end{center}
\vspace{-5mm}
\caption{ Quantitative results on DAVIS-585 benchmark. The metrics `NoC' and `NoF' mean the average Number of Clicks required and the Number of Failure examples for the target IOU. All models are trained on COCO~\cite{lin2014coco}+LVIS~\cite{gupta2019lvis}.  }
\label{tab:interDAVIS}
\end{table*}

\subsection{ Performance for Mask Correction}
We evaluate the performance of mask correction on the DAVIS-585 benchmark, and the results are listed in Table~\ref{tab:interDAVIS}. We also report the results of annotating from scratch. 

All initial masks provided by DAVIS-585 have IOUs between 0.75 to 0.85, and some challenging details have already been well annotated. Hence, making good use of them could logically facilitate the annotation. However, according to  Table~\ref{tab:interDAVIS}, RITM~\cite{sofiiuk2021ritm} do not show much difference between starting from initial masks and scratch.   In contrast, FocalClick makes good use of the initial masks. It requires significantly smaller numbers of clicks when starting from preexisting masks. Besides, it shows that the S1 version of FocalClick could outperform the big version of RITM~\cite{sofiiuk2021ritm} for mask correction tasks with 1/67 FLOPs.

\subsection{ Ablation Study}
We conduct plenty of ablation studies for our new modules and report experimental results on both the original DAVIS  and our DAVIS-585 dataset. \\

\begin{table}[t]
\small
\begin{center}
\scalebox{0.75}{
\begin{tabular}{ll|c|c|c|c|c|c}
\toprule[1pt]
\multicolumn{2}{l|}{} & \multicolumn{3}{c|}{DAVIS} & \multicolumn{3}{c}{DAVIS-585}   \\
\cline{3-8}

\multicolumn{2}{l|}{Method } & NoC85 & NoC90 &  NoF90 & NoC85 & NoC90  & NoF90 \\
\hline

\multicolumn{2}{l|}{Naive-B0-S1} &  10.70 & 15.60 & 250 & 12.26 & 15.99 & 441  \\
\multicolumn{2}{l|}{+ TC} &  5.70 & 9.56 & 119 & 5.84 & 9.45 & 184  \\
\multicolumn{2}{l|}{+ TC+ FC} &  5.15 & 7.66 & 72 & 5.41 & 8.52 & 145  \\
\multicolumn{2}{l|}{+ TC+ FC+ PM} &  5.13 & 7.42 & 64 & 2.63 & 3.69 & 54  \\
\hline
\multicolumn{2}{l|}{Naive-B0-S2} &  5.24 & 9.74 & 129 & 7.00 & 10.81 & 251  \\
\multicolumn{2}{l|}{+ TC} &  4.52 & 5.86 & 58 & 4.02 & 6.53 & 99 \\
\multicolumn{2}{l|}{+ TC+ FC} &  4.15 & 5.55 & 56 & 3.94 & 6.23 & 93  \\
\multicolumn{2}{l|}{+ TC+ FC+ PM} &  4.04 & 5.49 & 55 & 2.21 & 3.08 & 41  \\
\bottomrule[1pt]
\end{tabular}}
\end{center}
\vspace{-5mm}
\caption{ Ablation studies on both interactive segmentation from scratch and interactive mask correction. `TC', `FC', `PM' denote Target Crop, Focus Crop, and Progressive Merge. `NoC', `NoF' stand for the Number of Clicks and the Number of Failures. }
\label{tab:ablation}
\end{table}

\noindent \textbf{Holistic analysis.~}We verify the effectiveness for each of our novel components in Table~\ref{tab:ablation}.  We first construct a naive baseline model based on SegFormer-B0, noted as Naive-B0-S1/S2. It takes the full image as input and does not apply TC~(Target Crop), FC~(Focus Crop), and PM~(Progressive Merge), which is similar to early works like \cite{xu2016deep,jang2019brs,li2018latentdiversity}.  It is shown that this kind of pipeline performs poorly, especially for small input resolutions S1~($128\times128$). Most test samples fail to reach the target IOU within 20 clicks.  Then, after progressively adding the TC, FC, and PM, we observe that each component brings steady improvement for annotating from both initial masks and scratch. 

Making comparisons between S1 and S2, we find that the naive version heavily relies on the input scale. The performance drops tremendously from S2 to S1. However, with the assistance of TC, FC, and PM, the disadvantage of small input could be compensated. \\

\noindent \textbf{Analysis for cropping strategy.~}We first count the average area of the Focus Crop, Target Crop and calculate the ratio to full image in Table~\ref{tab:area}. It shows that our cropping strategy is effective in selecting and zoom-in the local regions.

\begin{table}[t]
\small
\begin{center}
\scalebox{0.75}{
\begin{threeparttable}
\begin{tabular}{l|c|c|c|c|c  }
\toprule[1pt]
    Ratio to Image    &  GrabCut &  Berkeley &  SBD  & DAVIS  & DAVIS-585 \\
\hline
Focus Crop  & 54.15\% & 31.17\% & 10.15\% & 11.6\% & 8.76\%\\
Target Crop  & 89.34\% & 68.81\% & 27.56\% & 40.50\% & 28.93\%\\
\bottomrule[1pt]
\end{tabular}
\end{threeparttable}
}
\end{center}
\vspace{-5mm}
\caption{ Statistics for the area of Focus Crop and Target Crop. We report the ratio relative to the full scale image.
}
\vspace{-2mm} 
\label{tab:area}
\end{table}

In Table~\ref{tab:refiner}, we verify the robustness of our cropping strategy. The results show that the fluctuation caused by the hyper-parameter is negligible compared with the improvement brought by the modules. Besides, for the evaluation result in Table.~\ref{tab:evaluation sota}, \ref{tab:interDAVIS},  we simply set those ratios as 1.4 following previous works~\cite{fbrs,sofiiuk2021ritm,chen2021cdnet}. However, Table~\ref{tab:refiner} shows that our work could reach even higher performance with a more delicate tuning.

\begin{figure}[h]
\newcommand{\image}
{\includegraphics[height=2cm]}
\centering 
\tabcolsep=0.07cm
\renewcommand{\arraystretch}{0.06}
\begin{tabular}{ccc}
\vspace{1mm}
\image{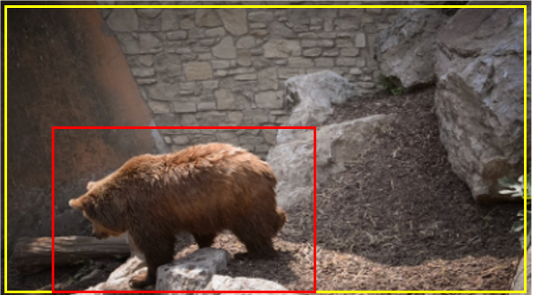} &
\image{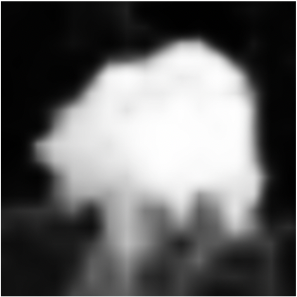} &
\image{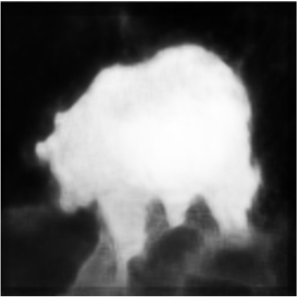} \\
\vspace{1mm}
\image{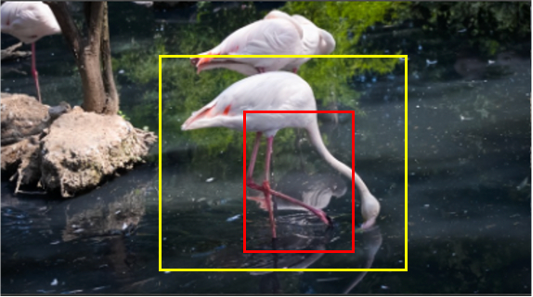} &
\image{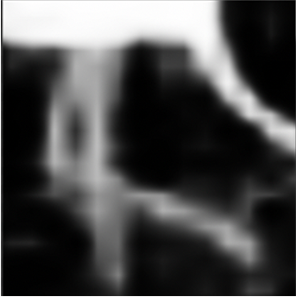} &
\image{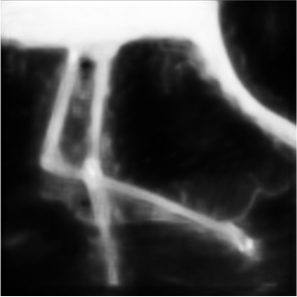} \\
\vspace{1mm}
\end{tabular}
\vspace{-3mm}
\caption{ Qualitative results for the effectiveness of Refiner. The first column denotes the Target Crop in yellow and the Focus Crop in red. The second and the third column demonstrate the mask in focus crop before and after refinement.
}
\label{fig:refiner}
\vspace{-6mm}
\end{figure}

We also visualize the intermediate results of the Refiner to demonstrate its effectiveness. In Fig.~\ref{fig:refiner}, the red boxes in the first column show the region selected by Focus Crop. The yellow box denotes the Target Crop~(The first row shows the case of the first click; hence the Target Crop corresponds to the entire image).
The second and the third column show the prediction results of the Segmentor and the Refiner. It demonstrates that Refiner is crucial for recovering the fine details. 

\begin{table}[t]
\begin{center}
\scalebox{0.60}{
\begin{threeparttable}
\begin{tabular}{l|c|c|c|c|c  }
\toprule[1pt]
\diagbox{ratio\_{FC}}{ratio\_{TC}}        &  1.2 &  1.4 &  1.6  & 1.8  & w/o TC \\
\hline
1.2  & \cellcolor{lightpink}5.15/7.21 & \cellcolor{lightpink}5.16/7.50 & \cellcolor{lightpink}5.23/7.90 & \cellcolor{lightpink}5.37/8.21 & 6.37/10.49\\
1.4  & \cellcolor{lightpink}5.07/7.10  & \cellcolor{lightpink}5.13/7.42 & \cellcolor{lightpink}5.15/7.77 & \cellcolor{lightpink}5.30/8.23 & 6.19/10.32\\
1.6  & \cellcolor{lightpink}4.99/7.07 & \cellcolor{lightpink}5.10/7.31 & \cellcolor{lightpink}5.11/7.80 & \cellcolor{lightpink}5.20/8.17 & 6.26/10.25\\
1.8  & \cellcolor{lightpink}4.99/6.95 & \cellcolor{lightpink}5.07/7.33 & \cellcolor{lightpink}5.08/7.64 & \cellcolor{lightpink}5.28/7.96 & 6.26/10.19\\
w/o FC  & 5.56/9.03 & 5.70/9.56 & 6.01/10.47 &6.56/11.31 &  \cellcolor{darkseagreen} 10.70/15.60\\
\bottomrule[1pt]
\end{tabular}
\end{threeparttable}
}
\end{center}
\vspace{-3mm}
\caption{ Combinations of the expand ratio for TC~(Target Crop) and FC~(Focus Crop). The values show the NoC80/90 on DAVIS. The last row/column shows the performance without FC/TC.
}
\vspace{-3mm}
\label{tab:refiner}
\end{table}

\begin{figure*}[t]
\newcommand{\image}{\includegraphics[width=0.49\columnwidth]}
\centering 
\tabcolsep=0.05cm
\renewcommand{\arraystretch}{0.06}
\begin{tabular}{cccc}
\vspace{0.5mm}
\image{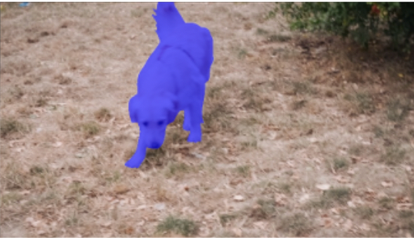} &
\image{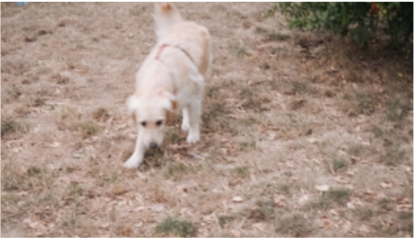} &
\image{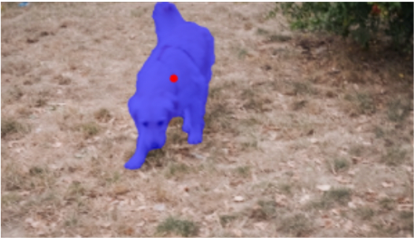} &
\image{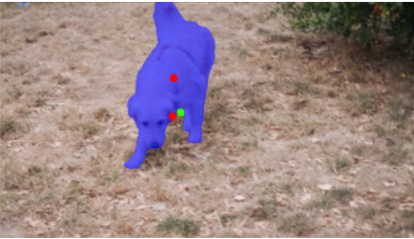} \\
\vspace{1mm}
{\footnotesize (1)~Ground Truth} & {\footnotesize Initial IOU : 0 } &{\footnotesize 1-Click : 92.1\% }  & {\footnotesize 3-Click IOU : 95.2\% } \\

\vspace{0.5mm}
\image{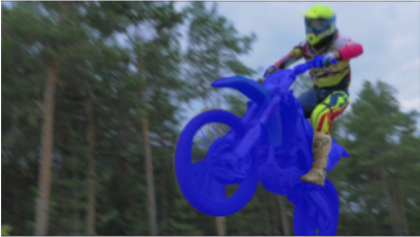} &
\image{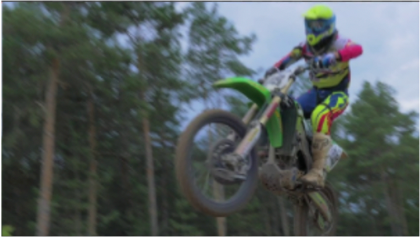} &
\image{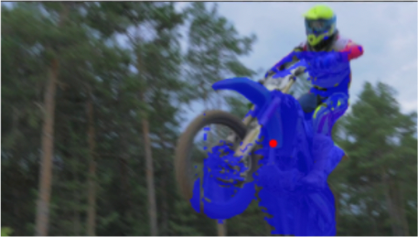} &
\image{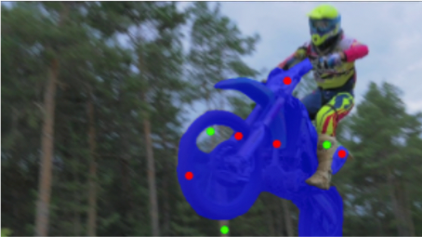} \\
\vspace{1mm}
{\footnotesize (2)~Ground Truth} & {\footnotesize Initial IOU : 0 } &{\footnotesize 1-Click : 52.7\% }  & {\footnotesize 8-Click IOU : 91.3\% } \\

\vspace{0.5mm}
\image{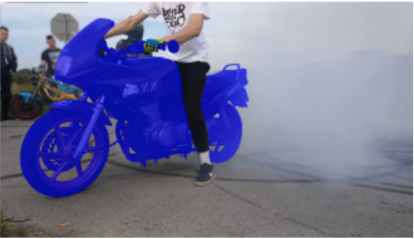} &
\image{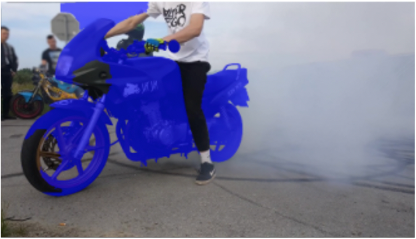} &
\image{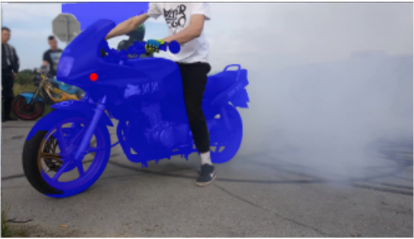} &
\image{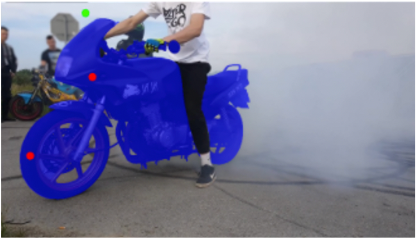} \\
\vspace{1mm}
{\footnotesize (3)~Ground Truth} & {\footnotesize Initial IOU : 81.3\% } &{\footnotesize 1-Click : 85.9\% }  & {\footnotesize 3-Click IOU : 98.5\% } \\

\vspace{0.5mm}
\image{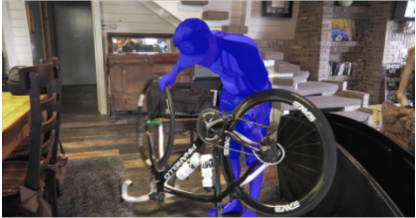} &
\image{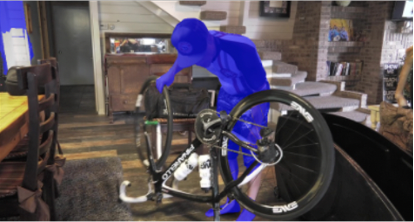} &
\image{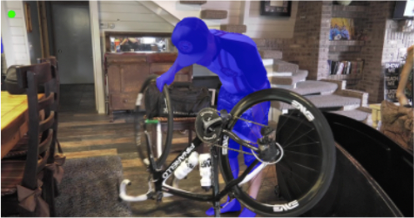} &
\image{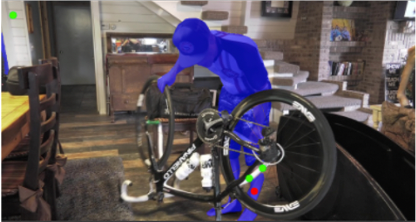} \\
\vspace{1mm}
{\footnotesize (4)~Ground Truth} & {\footnotesize Initial IOU : 83.8\% } &{\footnotesize 1-Click : 92.1\% }  & {\footnotesize 4-Click IOU : 95.2\% } \\

\vspace{0.5mm}
\image{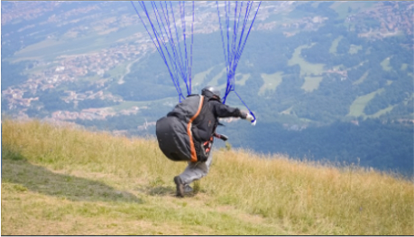} &
\image{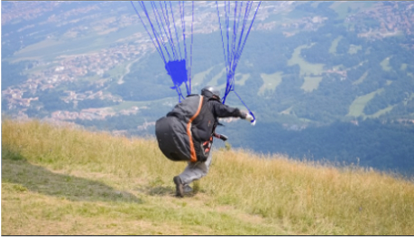} &
\image{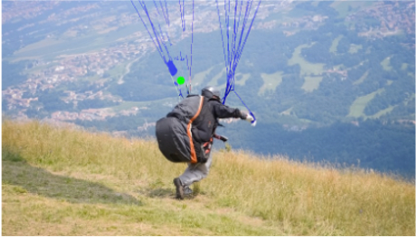} &
\image{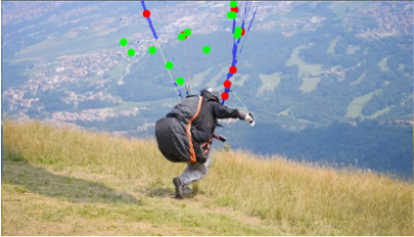} \\
\vspace{1mm}
{\footnotesize (5)~Ground Truth} & {\footnotesize Initial IOU : 75.6\% } &{\footnotesize 1-Click : 70.1\% }  & {\footnotesize 20-Click IOU : 23.7\% } \\
\end{tabular}
\vspace{-3mm}
\caption{Qualitative results on DAVIS-585 benchmarks. The results are predicted by FocalClick-B3-S2. Row 1, 2 show the example of annotation from scratch. Row 3, 4 demonstrate the cases starting from initial masks. Row 5 shows the bad cases.  \\}
\vspace{-7mm}
\label{fig:demo}
\end{figure*}

\subsection{Qualitative Result}
Qualitative results are demonstrated in Fig.~\ref{fig:demo}. The first two rows demonstrate the annotation process from scratch. FocalClick gives high-quality predictions within several clicks. The last three rows show the cases starting from preexisting masks. Our method completely preserves the well-segmented details and updates the regions that require correction. 
Row 5 gives a failure case, showing that FocalClick cannot annotate tiny structures like the parachute rope. In this case, the annotator could zoom in and switch to the brush tool to modify it manually. FocalClick supports users to switch back and continue the intelligent annotation for other regions.   
\vspace{-2mm}

\section{Limitation}
FocalClick improves the efficiency and the compatibility of existing pipelines. However, it still has the following limitations: 1)~As the failure cases in Fig.~\ref{fig:demo}, the performance on tiny structures is still not satisfactory. It could be further improved by leveraging more finely annotated data or matting datasets. 2)~For images under 1080P, the time of image loading, moving, zoom-in, and visualization can be ignored. However, for the 4K image, they would be the new bottleneck for speed. To build a practical annotation system, extensive engineering efforts are also required. 
\vspace{-3mm}

\section{Conclusion}
In this paper, we propose FocalClick to solve the practical problems for interactive segmentation. FocalClick significantly improves the efficiency of the existing pipeline, making it possible to be deployed on low-power devices. We also formulate a new task of interactive mask correction to meet the real-world requirements and propose corresponding solutions.

\noindent\textbf{Acknowledgement.} This work is supported in part by HKU Startup Fund.
{\small
\bibliographystyle{ieee_fullname}
\bibliography{egbib}

\begin{thebibliography}{10}\itemsep=-1pt

\bibitem{bai2014error}
Junjie Bai and Xiaodong Wu.
\newblock Error-tolerant scribbles based interactive image segmentation.
\newblock In {\em CVPR}, 2014.

\bibitem{borji2015msra10k}
Ali Borji, Ming-Ming Cheng, Huaizu Jiang, and Jia Li.
\newblock Salient object detection: A benchmark.
\newblock {\em TIP}, 2015.

\bibitem{boykov2001interactive}
Yuri~Y Boykov and M-P Jolly.
\newblock Interactive graph cuts for optimal boundary \& region segmentation of
  objects in nd images.
\newblock In {\em ICCV}, 2001.

\bibitem{chen2017deeplab}
Liang-Chieh Chen, George Papandreou, Iasonas Kokkinos, Kevin Murphy, and Alan~L
  Yuille.
\newblock Deeplab: Semantic image segmentation with deep convolutional nets,
  atrous convolution, and fully connected crfs.
\newblock {\em TPAMI}, 2017.

\bibitem{chen2021cdnet}
Xi Chen, Zhiyan Zhao, Feiwu Yu, Yilei Zhang, and Manni Duan.
\newblock Conditional diffusion for interactive segmentation.
\newblock In {\em ICCV}, 2021.

\bibitem{chollet2017xception}
Fran{\c{c}}ois Chollet.
\newblock Xception: Deep learning with depthwise separable convolutions.
\newblock In {\em CVPR}, 2017.

\bibitem{cong2020dovenet}
Wenyan Cong, Jianfu Zhang, Li Niu, Liu Liu, Zhixin Ling, Weiyuan Li, and Liqing
  Zhang.
\newblock Dovenet: Deep image harmonization via domain verification.
\newblock In {\em CVPR}, 2020.

\bibitem{dai2014synthesizability}
Dengxin Dai, Hayko Riemenschneider, and Luc Van~Gool.
\newblock The synthesizability of texture examples.
\newblock In {\em CVPR}, 2014.

\bibitem{everingham2010pascal}
Mark Everingham, Luc Van~Gool, Christopher~KI Williams, John Winn, and Andrew
  Zisserman.
\newblock The pascal visual object classes (voc) challenge.
\newblock {\em IJCV}, (2), 2010.

\bibitem{forte2020getting99}
Marco Forte, Brian Price, Scott Cohen, Ning Xu, and Fran{\c{c}}ois Piti{\'e}.
\newblock Getting to 99\% accuracy in interactive segmentation.
\newblock {\em arXiv:2003.07932}, 2020.

\bibitem{grady2006random}
Leo Grady.
\newblock Random walks for image segmentation.
\newblock {\em TPAMI}, 2006.

\bibitem{gulshan2010geodesic}
Varun Gulshan, Carsten Rother, Antonio Criminisi, Andrew Blake, and Andrew
  Zisserman.
\newblock Geodesic star convexity for interactive image segmentation.
\newblock In {\em CVPR}, 2010.

\bibitem{gupta2019lvis}
Agrim Gupta, Piotr Dollar, and Ross Girshick.
\newblock Lvis: A dataset for large vocabulary instance segmentation.
\newblock In {\em CVPR}, 2019.

\bibitem{hao2021edgeflow}
Yuying Hao, Yi Liu, Zewu Wu, Lin Han, Yizhou Chen, Guowei Chen, Lutao Chu,
  Shiyu Tang, Zhiliang Yu, Zeyu Chen, et~al.
\newblock Edgeflow: Achieving practical interactive segmentation with
  edge-guided flow.
\newblock In {\em ICCV}, 2021.

\bibitem{SBD}
Bharath Hariharan, Pablo Arbel{\'a}ez, Lubomir Bourdev, Subhransu Maji, and
  Jitendra Malik.
\newblock Semantic contours from inverse detectors.
\newblock In {\em ICCV}. IEEE, 2011.

\bibitem{he2017mask}
Kaiming He, Georgia Gkioxari, Piotr Doll{\'a}r, and Ross Girshick.
\newblock Mask r-cnn.
\newblock In {\em ICCV}, 2017.

\bibitem{howard2017mobilenets}
Andrew~G Howard, Menglong Zhu, Bo Chen, Dmitry Kalenichenko, Weijun Wang,
  Tobias Weyand, Marco Andreetto, and Hartwig Adam.
\newblock Mobilenets: Efficient convolutional neural networks for mobile vision
  applications.
\newblock {\em arXiv:1704.04861}, 2017.

\bibitem{jang2019brs}
Won-Dong Jang and Chang-Su Kim.
\newblock Interactive image segmentation via backpropagating refinement scheme.
\newblock In {\em CVPR}, 2019.

\bibitem{kim2010nonparametric}
Tae~Hoon Kim, Kyoung~Mu Lee, and Sang~Uk Lee.
\newblock Nonparametric higher-order learning for interactive segmentation.
\newblock In {\em CVPR}, 2010.

\bibitem{lempitsky2009image}
Victor Lempitsky, Pushmeet Kohli, Carsten Rother, and Toby Sharp.
\newblock Image segmentation with a bounding box prior.
\newblock In {\em ICCV}, 2009.

\bibitem{li2004lazy}
Yin Li, Jian Sun, Chi-Keung Tang, and Heung-Yeung Shum.
\newblock Lazy snapping.
\newblock {\em TOG}, 2004.

\bibitem{li2018latentdiversity}
Zhuwen Li, Qifeng Chen, and Vladlen Koltun.
\newblock Interactive image segmentation with latent diversity.
\newblock In {\em CVPR}, 2018.

\bibitem{liew2017regional}
JunHao Liew, Yunchao Wei, Wei Xiong, Sim-Heng Ong, and Jiashi Feng.
\newblock Regional interactive image segmentation networks.
\newblock In {\em ICCV}, 2017.

\bibitem{thinobject}
Jun~Hao Liew, Scott Cohen, Brian Price, Long Mai, and Jiashi Feng.
\newblock Deep interactive thin object selection.
\newblock In {\em WACV}, 2021.

\bibitem{liew2019multiseg}
Jun~Hao Liew, Scott Cohen, Brian Price, Long Mai, Sim-Heng Ong, and Jiashi
  Feng.
\newblock Multiseg: Semantically meaningful, scale-diverse segmentations from
  minimal user input.
\newblock In {\em ICCV}, 2019.

\bibitem{lin2014coco}
Tsung-Yi Lin, Michael Maire, Serge Belongie, James Hays, Pietro Perona, Deva
  Ramanan, Piotr Doll{\'a}r, and C~Lawrence Zitnick.
\newblock Microsoft coco: Common objects in context.
\newblock In {\em ECCV}, 2014.

\bibitem{firstclick}
Zheng Lin, Zhao Zhang, Lin-Zhuo Chen, Ming-Ming Cheng, and Shao-Ping Lu.
\newblock Interactive image segmentation with first click attention.
\newblock In {\em CVPR}, 2020.

\bibitem{long2015fcn}
Jonathan Long, Evan Shelhamer, and Trevor Darrell.
\newblock Fully convolutional networks for semantic segmentation.
\newblock In {\em CVPR}, 2015.

\bibitem{mahadevan2018iteratively}
Sabarinath Mahadevan, Paul Voigtlaender, and Bastian Leibe.
\newblock Iteratively trained interactive segmentation.
\newblock In {\em BMVC}, 2018.

\bibitem{majumder2019content}
Soumajit Majumder and Angela Yao.
\newblock Content-aware multi-level guidance for interactive instance
  segmentation.
\newblock In {\em CVPR}, 2019.

\bibitem{berkeley}
Kevin McGuinness and Noel~E O’connor.
\newblock A comparative evaluation of interactive segmentation algorithms.
\newblock {\em Pattern Recognition}, 2010.

\bibitem{peng2017largekernel}
Chao Peng, Xiangyu Zhang, Gang Yu, Guiming Luo, and Jian Sun.
\newblock Large kernel matters--improve semantic segmentation by global
  convolutional network.
\newblock In {\em CVPR}, 2017.

\bibitem{davis}
Federico Perazzi, Jordi Pont-Tuset, Brian McWilliams, Luc Van~Gool, Markus
  Gross, and Alexander Sorkine-Hornung.
\newblock A benchmark dataset and evaluation methodology for video object
  segmentation.
\newblock In {\em CVPR}, 2016.

\bibitem{rother2004grabcut}
Carsten Rother, Vladimir Kolmogorov, and Andrew Blake.
\newblock " grabcut" interactive foreground extraction using iterated graph
  cuts.
\newblock {\em TOG}, 2004.

\bibitem{fbrs}
Konstantin Sofiiuk, Ilia Petrov, Olga Barinova, and Anton Konushin.
\newblock f-brs: Rethinking backpropagating refinement for interactive
  segmentation.
\newblock In {\em CVPR}, 2020.

\bibitem{sofiiuk2021ritm}
Konstantin Sofiiuk, Ilia~A Petrov, and Anton Konushin.
\newblock Reviving iterative training with mask guidance for interactive
  segmentation.
\newblock {\em arXiv:2102.06583}, 2021.

\bibitem{tan2019efficientnet}
Mingxing Tan and Quoc Le.
\newblock Efficientnet: Rethinking model scaling for convolutional neural
  networks.
\newblock In {\em ICML}, 2019.

\bibitem{wang2020hrnet}
Jingdong Wang, Ke Sun, Tianheng Cheng, Borui Jiang, Chaorui Deng, Yang Zhao,
  Dong Liu, Yadong Mu, Mingkui Tan, Xinggang Wang, et~al.
\newblock Deep high-resolution representation learning for visual recognition.
\newblock {\em TPAMI}, 2020.

\bibitem{wang2017learningduts}
Lijun Wang, Huchuan Lu, Yifan Wang, Mengyang Feng, Dong Wang, Baocai Yin, and
  Xiang Ruan.
\newblock Learning to detect salient objects with image-level supervision.
\newblock In {\em CVPR}, 2017.

\bibitem{wu2014milcut}
Jiajun Wu, Yibiao Zhao, Jun-Yan Zhu, Siwei Luo, and Zhuowen Tu.
\newblock Milcut: A sweeping line multiple instance learning paradigm for
  interactive image segmentation.
\newblock In {\em CVPR}, 2014.

\bibitem{xiao2018upernet}
Tete Xiao, Yingcheng Liu, Bolei Zhou, Yuning Jiang, and Jian Sun.
\newblock Unified perceptual parsing for scene understanding.
\newblock In {\em ECCV}, 2018.

\bibitem{xie2021segformer}
Enze Xie, Wenhai Wang, Zhiding Yu, Anima Anandkumar, Jose~M Alvarez, and Ping
  Luo.
\newblock Segformer: Simple and efficient design for semantic segmentation with
  transformers.
\newblock In {\em NeurIPS}, 2021.

\bibitem{xu2017dim}
Ning Xu, Brian Price, Scott Cohen, and Thomas Huang.
\newblock Deep image matting.
\newblock In {\em CVPR}, 2017.

\bibitem{xu2016deep}
Ning Xu, Brian Price, Scott Cohen, Jimei Yang, and Thomas~S Huang.
\newblock Deep interactive object selection.
\newblock In {\em CVPR}, 2016.

\bibitem{xu2018youtubevos}
Ning Xu, Linjie Yang, Yuchen Fan, Dingcheng Yue, Yuchen Liang, Jianchao Yang,
  and Thomas Huang.
\newblock Youtube-vos: A large-scale video object segmentation benchmark.
\newblock {\em arXiv preprint arXiv:1809.03327}, 2018.

\bibitem{yuan2020ocr}
Yuhui Yuan, Xilin Chen, and Jingdong Wang.
\newblock Object-contextual representations for semantic segmentation.
\newblock In {\em ECCV}, 2020.

\bibitem{pspnet}
Hengshuang Zhao, Jianping Shi, Xiaojuan Qi, Xiaogang Wang, and Jiaya Jia.
\newblock Pyramid scene parsing network.
\newblock In {\em CVPR}, 2017.

\bibitem{ade20k}
Bolei Zhou, Hang Zhao, Xavier Puig, Sanja Fidler, Adela Barriuso, and Antonio
  Torralba.
\newblock Scene parsing through ade20k dataset.
\newblock In {\em CVPR}, 2017.

\end{thebibliography}
}

\end{document}